\documentclass[11pt,a4paper]{article}
\usepackage[hyperref]{naaclhlt2019}
\usepackage{times}
\usepackage{latexsym}
\usepackage{amsmath}
\usepackage{amsfonts}
\usepackage{url}

\aclfinalcopy % Uncomment this line for the final submission
\def\aclpaperid{110} %  Enter the acl Paper ID here

\usepackage{multirow}
\usepackage{tikz-qtree}
\usepackage{tikz-dependency}
\usepackage{graphicx}
\usepackage{stfloats}
\graphicspath{{images/}}
\newcommand*{\LargerCdot}{\raisebox{-1.0ex}{\scalebox{3.0}{$\cdot$}}}
\usepackage{pgfplots}

\newcommand*\samethanks[1][\value{footnote}]{\footnotemark[#1]}

\definecolor{darkgreen}{HTML}{228B22}

\title{Syntax-aware Neural Semantic Role Labeling with Supertags}

\author{Jungo Kasai\textsuperscript{$\clubsuit$}\thanks{\ \ Work partially done at Yale University.} \quad\quad\quad Dan Friedman\textsuperscript{$\spadesuit$}\samethanks \quad\quad\quad Robert Frank\textsuperscript{$\diamondsuit$} \quad \quad \quad \\
{\bf Dragomir Radev\textsuperscript{$\diamondsuit$}} \quad\quad\quad {\bf Owen Rambow\textsuperscript{$\heartsuit$}} \quad\quad\quad\\
\textsuperscript{$\clubsuit$}University of Washington  \quad
\textsuperscript{$\spadesuit$}Google\\
\textsuperscript{$\diamondsuit$}Yale University \quad 
\textsuperscript{$\heartsuit$}Elemental Cognition, LLP\\
\scalebox{0.85}[0.9]{{\tt jkasai@cs.washington.edu}} \quad
\scalebox{0.85}[0.9]{{\tt danfriedman@google.com}}\\
\scalebox{0.85}[0.9]{{\tt \{robert.frank,dragomir.radev\}@yale.edu}} \quad
\scalebox{0.85}[0.9]{{\tt owenr@elementalcognition.com}}}

\date{}

\begin{document}
\maketitle
\begin{abstract}
  We introduce a new syntax-aware model for dependency-based semantic role labeling that outperforms syntax-agnostic models for English and Spanish. We use a BiLSTM to tag the text with supertags extracted from dependency parses, and we feed these supertags, along with words and parts of speech, into a deep highway BiLSTM for semantic role labeling. Our model combines the strengths of earlier models that performed SRL on the basis of a full dependency parse with more recent  models that use no syntactic information at all. Our local and non-ensemble model achieves state-of-the-art performance on the CoNLL 09 English and Spanish datasets. SRL models benefit from syntactic information, and we show that supertagging is a simple, powerful, and robust way to incorporate syntax into a neural SRL system.
\end{abstract}

\section{Introduction}
Semantic role labeling (SRL) is the task of identifying the semantic relationships between each predicate in a sentence and its arguments \cite{gildea2002}. %For example, in the sentence ``The investors \textbf{clashed} with the board over voting power,'' an SRL system should determine that the investors are ones who clashed, the board is the thing that was clashed with, and voting power is the thing that was clashed about. SRL information has been shown to be useful for such downstream tasks as machine translation \cite{bazrafshan2013semantic}, multi-document summarization \cite{yan2014srrank}, information extraction \cite{christensen2011analysis}, and question answering \cite{shen2007using}.
%Until recently it was believed that SRL systems require syntactic information to perform well \cite{punyakanok2008importance}.
%For example, in the work of
%\newcite{roth2016}, the output of a syntactic parser is given as input to a neural SRL model. 
While early research assumed that SRL models required syntactic information to perform well \cite{punyakanok2008importance}, recent work has demonstrated that neural networks can achieve competitive and even state-of-the-art performance without any syntactic information at all \cite{zhou2015,marcheggiani2017,he2017}. These systems have the benefits of being simpler to implement and performing more robustly on foreign languages and out-of-domain data, cases where syntactic parsing is more difficult \cite{marcheggiani2017}.

In this paper, we show that using supertags is an effective middle ground between using full syntactic parses and using no syntactic information at all. A supertag is a linguistically rich description assigned to a lexical item. Supertags impose complex constraints on their local context, so supertagging can be thought of as ``almost parsing'' \cite{bangalore1999supertagging}. Supertagging has been shown to facilitate Tree-Adjoining Grammar (TAG) parsing \cite{mica,friedman-EtAl:2017:TAG+13,kasai2017tag,kasai-EtAl:2018:N18-1} and Combinatory Categorial Grammar (CCG) parsing \cite{Clark2007WideCoverageES,kummerfeld-EtAl:2010:ACL,lewis2016lstm,xu:2016:EMNLP2016}.

%\newcite{ouchi2014improving} extract supertags from a dependency-annotated corpus and show that supertagging improves performance on dependency parsing.

We propose that supertags can serve as a rich source of syntactic information for downstream tasks without the need for full syntactic parsing. Following \newcite{ouchi2014improving}, who used supertags to improve dependency parsing, we extract various forms of supertags from the dependency-annotated CoNNL 09 corpus. 
This contrasts with prior SRL work that uses TAG or CCG supertags  \cite{chenrambow2003,lewis-he-zettlemoyer:2015:EMNLP}.
We train a bidirectional LSTM (BiLSTM) to predict supertags and feed the predicted supertag embedding, along with word and predicted part-of-speech embeddings, to another BiLSTM for semantic role labeling. Predicted supertags are represented by real-valued vectors, contrasting with approaches based on syntactic paths \cite{roth2016,he-EtAl:2018:Long} and syntactic edges \cite{marcheggiani-titov:2017:EMNLP2017,strubell-EtAl:2018:EMNLP}.
%(GCNs: \citet{marcheggiani-titov:2017:EMNLP2017}, Self-Attention: \citet{strubell-EtAl:2018:EMNLP}). 
This way of incorporating information alleviates the issue of error propagation from parsing.
%we will demonstrate that supertags ameliorate role labeling of out-of-domain data by an even larger margin compared to in-domain data.

Supertagging has many advantages as part of a natural language processing pipeline. First, as a straightforward sequence-labeling task, the supertagging architecture is much simpler than comparable systems for structured parsing.
%Thus we can make use of black-box implementations of bidirectional LSTMs, and, as deep learning research advances, we can easily incorporate new improvements into our system and expect to see improvements in our results.
Second, it is simple to extract different forms of supertags from a dependency corpus to test different hypotheses about which kinds of syntactic information are most useful for downstream tasks.
Our results show that supertags, by encoding just enough information, can improve SRL performance even compared to systems that incorporate complete dependency parses.

\vspace{-2mm}
\section{Our Models}

\vspace{-2mm}
\subsection{Supertag Design}
We experiment with four supertag models, two from \newcite{ouchi2014improving}, one from \newcite{Nguyen_Nguyen-2016}, and one of our own design inspired by Tree Adjoining Grammar supertags \cite{bangalore1999supertagging}. Each model encodes a different set of attributes about the syntactic relationship between a word, its parent, and its dependents. Table \ref{stag:table} summarizes what information is expressed in each supertag model.

% \newcite{ouchi2014improving} explore the use of two models (which they label Model 1 and Model 2) for supertags, and show that both improve dependency parsing performance.
% \newcite{Nguyen_Nguyen-2016} propose an even simpler supertag model for Vietnamese dependency parsing. 
% In this paper, we provide a systematic view over four types of dependency-based supertags, and explore their efficacy with regards to SRL. 

% Table \ref{stag:table} summarizes what information is expressed in each supertag model. As \newcite{ouchi2014improving} note, designing useful supertags necessitates finding the balance between granularity and predictablity: overly fine-grained supertags would express detailed syntactic information at the expense of low supertagging accuracy. Setting the syntax-agnostic model in \newcite{marcheggiani2017} as a baseline, we will show experiments using all four supertag models in a later section.

% \begin{figure}
% \centering
% \begin{dependency}
% \begin{deptext}
% ROOT \& No \& , \& it \& was \& n't \& black \& Monday \& .\\
% \end{deptext}
% \depedge{5}{2}{DEP}
% \depedge{5}{3}{P}
% \depedge{5}{4}{SBJ}
% \depedge{1}{5}{ROOT}
% \depedge{5}{6}{ADV}
% \depedge{8}{7}{NAME}
% \depedge{5}{8}{PRD}
% \depedge{5}{9}{P}
% \end{dependency}
% \caption{A gold parse given in the CoNLL 2009 English evaluation set.}
% \label{dependency}
% \end{figure}
\begin{table}
\footnotesize
\centering
\begin{tabular}{ |c |cc| }
\hline
  Token & Model 1 & Model TAG\\\hline
      No  & DEP/R&DEP/R\\
      ,  & P/R& P/R\\
      it &SBJ/R&-\\
      was  &ROOT+L\_R&ROOT+SBJ/L\_PRD/R  \\
      n't &ADV/L&ADV/L \\
      black &NAME/R&NAME/R \\
      Monday &PRD/L+L&-\\
%      . &P/L&P/L&P/L&P/L\\
      \hline
\end{tabular}
\caption{Supertags for the sentence ``No, it wasn't black Monday.''}
\label{stag:ex}
\vspace{-0.3cm}
\end{table}
%\begin{table*}[b!]
%\footnotesize
%\centering
%\begin{tabular}{ |c |cccc| }
%\hline
%  Token & Model 0 & Model 1  & Model 2 & Model TAG\\\hline
%      No  & DEP/R & DEP/R& DEP/R&DEP/R\\
%      ,  & P/R & P/R& P/R& P/R\\
%      it &SBJ/R&SBJ/R&SBJ/R&-\\
%      was  & ROOT&ROOT+L\_R&ROOT+SBJ/L\_PRD/R&ROOT+SBJ/L\_PRD/R  \\
%      n't &ADV/L&ADV/L&ADV/L&ADV/L \\
%      black &NAME/R&NAME/R&NAME/R&NAME/R \\
%      Monday &PRD/L&PRD/L+L&PRD/L+L&-\\
%%      . &P/L&P/L&P/L&P/L\\
%      \hline
%\end{tabular}
%\caption{Supertags for the sentence ``No, it wasn't black Monday.''}
%\label{stag:ex}
%\end{table*}

\begin{table}
 \footnotesize
 \centering
 \begin{tabular}{|c c | c  c c c|}
 \hline
 Position & Feature & 0& 1& 2 & TAG \\ \hline 
  \multirow{2}{*}{Obligatory Parent}&Direction &\LargerCdot& \LargerCdot&\LargerCdot&\\
  & Relation &\LargerCdot&\LargerCdot&\LargerCdot&\\\hline
  \multirow{2}{*}{Optional Parent}&Direction &\LargerCdot& \LargerCdot&\LargerCdot &\LargerCdot\\\
  &Relation&\LargerCdot&\LargerCdot&\LargerCdot&\LargerCdot\\ \hline
 % &POS&&&&&&\LargerCdot&\LargerCdot\\ \hline
  \multirow{2}{*}{Obligatory Dep.}&Direction&&\LargerCdot&\LargerCdot&\LargerCdot\\
  &Relation&&&\LargerCdot&\LargerCdot\\ \hline
  \multirow{1}{*}{Optional Dep.}&Direction&&\LargerCdot&\LargerCdot&\\ \hline
  %&Relation&&&&&&&\\ \hline
 \end{tabular}

\caption{Supertag models for SRL. Models 1 and 2 are from \newcite{ouchi2014improving} and Model 0 is from \newcite{Nguyen_Nguyen-2016}.}
 \label{stag:table}
\end{table}

\noindent\textbf{Model 0.}
A Model 0 supertag for a word $w$ encodes the dependency relation and the relative position (direction) between $w$ and its  head, i.e. left (L), right (R), or no direction (ROOT) \cite{Nguyen_Nguyen-2016}.

\noindent\textbf{Model 1.}
A Model 1 supertag for $w$ adds to the ``parent information" from Model 0 the information of whether $w$ possesses dependents to its left (L) or right (R) \cite{ouchi2014improving}.
%Model 1

\noindent\textbf{Model 2.}
A Model 2 supertag  for $w$ extends Model 1 by encoding the dependency relation between  $w$ and its obligatory dependents.\footnote{Following \newcite{ouchi2014improving}, we define obligatory dependents as those with relations `SBJ,' `OBJ,' `PRD,' and `VC.' For Spanish, we define obligatory syntactic arguments as `dc,'`suj,' `cd,' and `cpred.'} When $w$ lacks such obligatory children, we encode whether it possesses non-obligatory dependents to the left (L) or right (R) as in Model 1.

\noindent\textbf{Model TAG.}
We propose Model TAG supertags that represent syntactic information analogously to TAG supertags (elementary trees) \cite{bangalore1999supertagging}. %Seen in Figure \ref{modeltag} are the TAG supertags for \textit{John really likes Mary}. 
%Notice that \textit{John} and \textit{Mary} get assigned the same supertag, since the supertag does not encode parent information, in contrast to Models 0-2. Like Models 1 and 2, %a TAG supertag for \textit{likes} represents the presence of the subject to the left and the object to the right. The supertag for \textit{really}, an auxiliary tree, on the other hand, tells us that the adjunct (optional modifier) modifies a verb to the right. We can extract such supertags analogously from dependency parses. 
A Model TAG supertag encodes the dependency relation and the direction of the head of a word similarly to Model 0 if the dependency relation is non-obligatory (corresponding to \textit{adjunction} nodes), and the information about obligatory dependents of verbs if any similarly to Model 2 (corresponding to \textit{substitution} nodes).

\vspace{-2mm}
%\footnote{We found that word segmentaiton in the CoNLL 2019 English dataset differed from PTB, yielding two POS tags (PRF and NIL) not present in the PTB tag set. We map these tags to a new coarse-grained POS tag.}
\subsection{Supertagger Model}
%Motivated by recent state-of-the-art supertaggers (e.g. TAG: \newcite{kasai2017tag}; CCG: \newcite{lewis2016lstm}; \newcite{xu:2016:EMNLP2016}), we employ a  bi-directional LSTM (BiLSTM) architecture (Figure \ref{stagger}) for our dependency-based supertagging. 

% cut the picture for bilstm stagger
Motivated by recent state-of-the-art supertaggers (TAG: \newcite{kasai2017tag,kasai-EtAl:2018:N18-1}; CCG: \newcite{lewis2016lstm}; \newcite{xu:2016:EMNLP2016}), we employ a  bi-directional LSTM (BiLSTM) architecture for our supertagging. 
%\begin{figure}[h] 
%\hspace{-4mm}
%    \centering
%    \includegraphics[width=0.45\textwidth]{stagger_new.png}
%    \caption{BiLSTM Supertagger Architecture.}
%    \label{stagger}
%%\vspace{-2mm}
%\end{figure}
The input for each
word is the conncatenation of a
dense vector representation of the word, a vector embedding of a predicted PTB-style POS tag (only for English),\footnote{For the English data, predicted PTB-style POS tags generally contribute to increases, approximately 0.2-0.4\% in the dev set, whereas for Spanish adding predicted (coarse-grained) POS tags hurt the performance.} and a vector output by character-level Convolutional Neural Networks (CNNs) for morphological information.

For POS tagging before English supertagging, we use the same hyperparameters as in \newcite{ma-hovy:2016:P16-1}. 
For supertagging, we follow the hyperparameters chosen in \newcite{kasai-EtAl:2018:N18-1} regardless of the supertag model that is employed.
%Specifically, we use two layers of BiLSTMs with 512 units each. Input, layer-to-layer, and recurrent \cite{Gal2016Theoretically} dropout rates are all 0.5.
%We embed predicted POS tags by randomly initialized 100 dimensional vectors.
We initialize the word embeddings by the pre-trained 100 dimensional GloVe \cite{pennington2014glove} and the 300 dimensional FastText \cite{Q17-1010} vectors for English and Spanish respectively.
%For training, we perform 5-fold jackknife training to obtain predicted supertags.
%Supertagging accuracies are shown in the appendix. 

% \begin{figure}
% \centering
% \begin{dependency}
% \begin{deptext}
% ROOT \& No \& , \& it \& was \& n't \& black \& Monday \& .\\
% \end{deptext}
% \depedge{5}{2}{DEP}
% \depedge{5}{3}{P}
% \depedge{5}{4}{SBJ}
% \depedge{1}{5}{ROOT}
% \depedge{5}{6}{ADV}
% \depedge{8}{7}{NAME}
% \depedge{5}{8}{PRD}
% \depedge{5}{9}{P}
% \end{dependency}
% \caption{A gold parse given in the CoNLL 2009 English evaluation set.}
% \label{dependency}
% \end{figure}

\vspace{-2mm}
\subsection{Semantic Role Labeling}
\begin{figure}[h] 
\hspace{-4mm}
    \centering
    \includegraphics[width=0.45\textwidth]{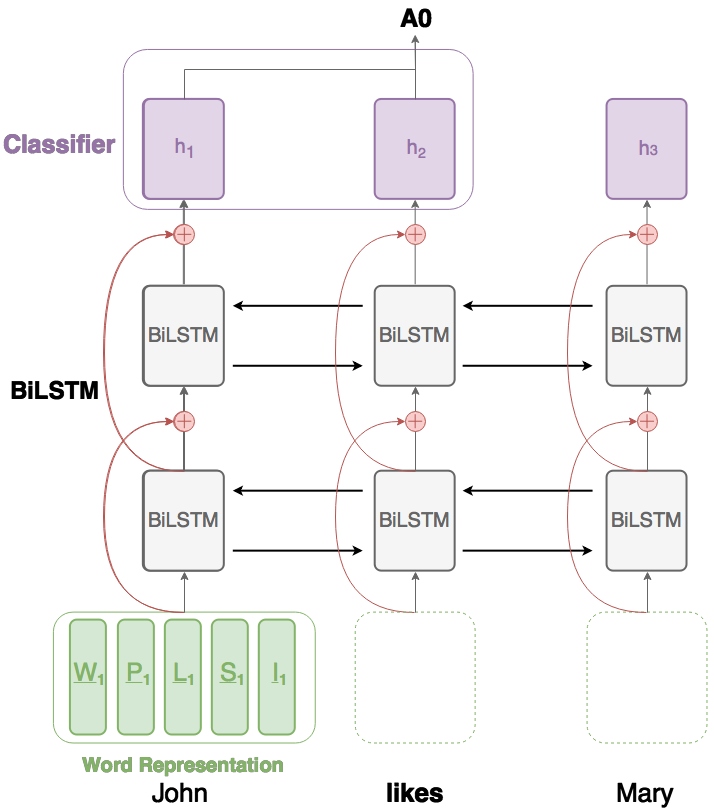}
    \caption{SRL architecture with a highway BiLSTM. $W_1$, $P_1$, $L_1$, $S_1$, $I_1$ indicate the word, POS, lemma, supertag, and predicate indicator embeddings for the first token, \textit{John}. Here we only show two layers.}
    \label{srl_architecture}
\vspace{-5mm}
\end{figure}

Our SRL model is most similar to the syntax-agnostic SRL model proposed by \newcite{marcheggiani2017}. Our model differs in two ways: 1) we add randomly initialized 50 dimensional supertag embeddings to the input layer (Fig.\ \ref{srl_architecture}), and 2) we use a modified LSTM with highway layers and regularization (0.5 dropout) as in \newcite{he2017}.

We use the same hyperparameters as in \newcite{marcheggiani2017} with randomly initialized 50 dimensional embeddings for supertags.\footnote{We provide lists of hyperparameters in Appedix \ref{hyper}.}
For pre-trained word embeddings, we use the same word embeddings as the ones in \newcite{marcheggiani2017} for English and the 300-dimensional FastText vectors \cite{Q17-1010} for Spanish.
We use the predicates predicted by the mate-tools \cite{bjorkelund-hafdell-nugues:2009:CoNLL-2009-ST} (English) and \newcite{zhao-EtAl:2009:CoNLL-2009-ST2} (Spanish) system in our models, again following \newcite{marcheggiani2017} to facilitate comparison.
Our code is available online for easy replication of our results.\footnote{\url{https://github.com/jungokasai/stagging_srl}.}
%\footnote{Original implementation: \url{https://github.com/jungokasai/stagging_srl}. Our reimplementation in AllenNLP is available at.}

\section{Results and Discussion}
%Table \ref{stag-result} provides our supertagging results for English and Spanish across the different types of supertag described above.
\begin{table*}
\footnotesize
\centering
\begin{tabular}{ |l |c c c c|c c c|}
\hline
  & \multicolumn{4}{c|}{English} & \multicolumn{3}{c|}{Spanish}\\
  Supertag &  \# Stags & Dev & ID & OOD & \# Stags &Dev &  ID \\\hline
%  Model 0&99&92.43&93.53&87.60&88&92.39&92.14&&&\\
  Model 0&99&92.93&94.17&88.71&88&92.97&92.67\\
  Model 1&298&91.07&92.50&86.51&220&90.63&90.37\\
%  Model 2&692&90.20&91.54&84.29&&&\\
  Model 2&692&90.60&92.05&85.40&503&90.08&89.84\\
%  Model TAG&430&92.14&93.33&86.33&&&\\\hline
  Model TAG&430&92.60&94.17&87.46&317&92.33&92.18\\\hline
%  Model 1-POS &1009&90.42&91.98&85.46&551&90.31&90.13\\
%  Model TAG-POS&1001&89.37&91.09&84.04&494&92.16&91.78\\\hline
%  Model 1 POS&1132&92.18&93.51&87.43&&91.74&91.53\\\hline
%  Model TAG POS &865&92.18&93.51&87.43&&91.74&91.53\\\hline
\end{tabular}
\caption{Supertagging accuracies for English and Spanish. ID and OOD indicate the in-domain and out-of-domain evaluation data respectively. The \# Stags columns show the number of supertags in the corresponding training set.}
\label{stag-result}
\vspace{-5mm}
\end{table*}

Table \ref{stag-result} provides our supertagging results for English and Spanish across the different types of supertag described above.
Here we clearly see the general pattern that the more granular supertagging becomes, the less reliable it is, and finding the balance between granularity and predictability is critical.
We present our SRL results in Tables \ref{srl1}-\ref{srl-es} along with the results from a baseline BiLSTM model, which is our implementation of the syntax-agnostic model in \newcite{marcheggiani2017}.
We also present results for a BiLSTM model with dropout and highway connections but without supertags (BDH model), to distinguish the effects of supertags from the effects of better LSTM regularization. In every experiment we train the model five times, and present the mean score. Table \ref{srl1} shows that Model 1 yields the best performance in the English dev set, and thus we only use Model 1 supertags for test evaluation.
We primarily show results only with word type embeddings to conduct fair comparisons with prior work, but we also provide results with deep contextual word representations, ELMo \cite{Peters2018}, and compare our results with recent work that utilizes ELMo \cite{he-EtAl:2018:Long}. \footnote{We used the pretrained ELMo available at \url{https://tfhub.dev/google/elmo/2}.}

\noindent\textbf{English in-domain.}
Table \ref{srl2} summarizes the results on the English in-domain test set. First, we were able to approximately replicate the results from \newcite{marcheggiani2017}.
Adding dropout and highway connections to our BiLSTM model improves performance by 0.5 points, to 88.1, and adding supertags improves results even further to 88.6.
Our supertag model performs even better than the non-ensemble model in \newcite{marcheggiani-titov:2017:EMNLP2017}, in which the model is given the complete dependency parse of the sentence. This result suggests that supertags can be even more effective for SRL than a more complete representation of syntax. 
Furthermore, our supertag-based method with contextual representations achieves 90.2, a new state-of-the-art.
Interestingly, the gain from supertagging decreases to 0.2 points (90.2 vs.\ 90.0) in the presence of contextual representations, suggesting that contextual representations encode some of the same syntactic information that supertags provide.
%But we cannot compare these models directly because our models gain additional improvements from dropout and highway connections.

\noindent\textbf{English out-of-domain.}
One of the advantages of using a syntax-agnostic SRL model is that such a model can perform relatively well on out-of-domain data, where the increased difficulty of syntactic parsing can cause errors in a syntax-based system \cite{marcheggiani2017}.
Unfortunately we were not able to replicate the out-of-domain results of \newcite{marcheggiani2017}: our implementation of the BiLSTM achieves a score of 76.4, compared to their reported score of 77.7. However, we note that incorporating supertags into our own model improves performance, 
with our best model achieving a score of 77.6.
%% ocr: was a significance test performed?  
%% jkasai: Let's just report s.d.
Our supertag-based model also substantially outperforms the full dependency-based models \cite{roth2016,marcheggiani-titov:2017:EMNLP2017}.
This suggests that syntax with a certain degree of granularity is useful even across domains.
Our supertag-based method alleviates the issue of error propagation from syntactic parsing.
Finally, our model with contextual representations yields 80.8, an improvement of 1.5 F1 points over the previous state-of-the-art \cite{he-EtAl:2018:Long}, which also uses ELMo.
%serving as a robust middle ground between syntax-agnostic and full parse-based approaches.

%\paragraph{Ensemble}
%Seen in Table \ref{srl-ensemble} are the results from ensembling of 3 networks with different initialization. Unlike the model in \newcite{marcheggiani-titov:2017:EMNLP2017}, we did not gain significant improvement in the test sets, falling in the range of the improvement in \newcite{roth2016} and \newcite{fitzgerald-EtAl:2015:EMNLP}. 
%
\noindent\textbf{Spanish.}
Table \ref{srl-es} shows the results on the Spanish test data. 
Our BiLSTM implementation yields lower performance than \newcite{marcheggiani2017}: our model achieves a score of 79.1, compared to their reported score of 80.3.
However, our BDH model yields a score of 80.8, already achieving state-of-the-art performance. 
Adding supertags to BDH improves the score further to 81.0.
This suggests that while the gains are relatively small, the supertag-based approach still helps Spanish SRL.
Supertags slightly improve performance when contextual representations are used (83.0 vs. 82.9).
See Appendix \ref{hyper} for details.

\begin{table}
\footnotesize
\centering
\begin{tabular}{ l |ccc }
\hline
  Architecture & P & R & F\textsubscript{1}\\\hline
      BiLSTM & 87.27&85.16&86.20 \\
      BiLSTM + DOut &86.49 & 86.11&86.30\\
      BiLSTM + DOut + HWay &86.97 &86.43&86.70 \\\hline
      BDH + Model 0 &87.47&86.46&86.96\\
      BDH + Model 1 &87.69&86.72&\textbf{87.20}\\
%      BDH + Model 1Gold &89.54&88.66&89.11\\
      BDH + Model 2 &87.54&86.09&86.81\\
      BDH + Model TAG &87.78 &86.07&86.92\\ 
%      BDH + Model 1 POS&87.70 &86.43&87.06\\
%      BDH + Model TAG-POS&87.51 &86.36&86.93\\\hline
%      BDH + Model X &&&\\\hline
\end{tabular}
\caption{Results on the CoNLL 2009 dev set for English. BDH stands for BiLSTM + Dropout + Highway.}
\label{srl1}
\vspace{-0.3cm}
\end{table}

\begin{table}
\footnotesize
\centering
\begin{tabular}{ l |ccc }
\hline
  \textbf{Non-ensemble System} & P & R & F\textsubscript{1}\\
      \newcite{fitzgerald-EtAl:2015:EMNLP}& --& --&87.3\\
      \newcite{roth2016}  & 90.0 &85.5 &87.7 \\
      \newcite{marcheggiani2017}&88.7& 86.8&87.7 \\
      \newcite{marcheggiani-titov:2017:EMNLP2017}& 89.1& 86.8&88.0 \\
      BiLSTM & 88.5&86.7&87.6\\
%      BiLSTM+DOut & 88.0&87.6&87.8\\
      BDH &88.3&87.8&88.1\\
%      BDH+Model 0 &88.8&88.0&88.4\\ old stags
%      BDH+Model 0 &88.8&88.1&88.5\\ 
      BDH + Model 1 &89.0&88.2&\textbf{88.6}\\
    \textbf{+ Contextual Representations} &  &  & \\
      \citet{he-EtAl:2018:Long} (ELMo) &89.7 &89.3 &89.5\\
%      BDH + ELMo &\\
      BDH + ELMo &90.3&89.7&90.0\\
      BDH + Model 1 + ELMo &90.3&90.0&\textbf{90.2}\\ \hline
  \textbf{Ensemble System} &&&\\
  \citet{fitzgerald-EtAl:2015:EMNLP} & --&--& 87.7\\
      \newcite{roth2016} &90.3&85.7&87.9\\
      \newcite{marcheggiani-titov:2017:EMNLP2017}& 90.5 & 87.7&89.1 \\\hline
%      BDH+Model 1 Gold &90.7&90.0&90.3\\
%      BDH+Model 2 &88.9&87.6&88.2\\ 
%      BDH+Model TAG &88.9&87.6&88.3\\  \hline
%      BDH+Model 1 POS &89.0&87.7&88.3\\
%      BDH+Model TAG-POS &88.8&87.8&88.3\\ \hline
%      BDH+Model X &&&88.2\\\hline\hline
%      Ensemble &&&\\
%      \newcite{roth2016} &90.3 &85.7& 87.9\\
%      \newcite{marcheggiani-titov:2017:EMNLP2017}&90.5 &87.7& 89.1\\
%      BDH &88.4&88.2&88.3\\
%      BDH+Model 0 &88.8&87.&88.3\\
%      BDH+Model 1 &89.5&87.7&88.6\\
%      BDH+Model 2 &&&\\
%      \hline
\end{tabular}
\caption{Results on the CoNLL 2009 in-domain test set for English. All standard deviations in $F_1 < 0.12$.}
\label{srl2}
%\vspace{-0.3cm}
\end{table}

%\begin{table}
%\small
%\centering
%\begin{tabular}{ c |ccc }
%\hline
%  Ensemble System & P & R & F\textsubscript{1}\\\hline\hline
%      \newcite{roth2016} &90.3 &85.7& 87.9\\
%      \newcite{marcheggiani-titov:2017:EMNLP2017}&90.5 &87.7& \textbf{89.1}\\\hline
%      BDH &88.4&88.2&88.3\\
%      BDH+Model 0 &88.8&87.8&88.3\\
%      BDH+Model 1 &89.5&87.8&88.6\\
%      BDH+Model 2 &88.9&87.6&88.3\\
%      BDH+Model TAG &88.4&88.4&88.4\\
%%      BDH+Model 1 POS &89.0&87.8&88.4\\
%%      BDH+Model TAG-POS &88.8&87.8&88.3\\
%      \hline
%\end{tabular}
%\caption{Ensemble system results on the CoNLL-2009 in-domain test set for English. The \newcite{roth2016} and \newcite{marcheggiani-titov:2017:EMNLP2017} are both ensemble models, which yield the best performance.}
%\end{table}

\begin{table}
\footnotesize
\centering
\begin{tabular}{ l |ccc }
\hline
  \textbf{Non-ensemble System} & P & R & F\textsubscript{1}\\
      \newcite{fitzgerald-EtAl:2015:EMNLP} & --& -- & 75.2 \\
      \newcite{roth2016}  & 76.9& 73.8 &75.3  \\
      \newcite{marcheggiani2017}&79.4&76.2 &\textbf{77.7}\\
      \newcite{marcheggiani-titov:2017:EMNLP2017}&78.5& 75.9& 77.2 \\
            BiLSTM & 77.2&75.6&76.4\\
%      BiLSTM+DOut &76.6 &76.1&76.3\\
      BDH &77.8&76.6&77.2\\
%      BDH+Model 0 &76.7&75.9&76.3\\ old stags
%      BDH+Model 0 &77.4&76.3&76.8\\
      BDH + Model 1 &78.0&77.2&77.6\\
  \textbf{+ Contextual Representations} &  &  & \\
      \citet{he-EtAl:2018:Long} (ELMo) &81.9&76.9&79.3\\
      BDH + ELMo &81.1&80.4&\textbf{80.8}\\
      BDH + Model 1 + ELMo &81.0&80.5&\textbf{80.8}\\\hline
  \textbf{Ensemble System} &  &  & \\
  \citet{fitzgerald-EtAl:2015:EMNLP} & --&--& 75.5\\
      \newcite{roth2016} &79.7&73.6&76.5\\
      \newcite{marcheggiani-titov:2017:EMNLP2017}& 80.8&77.1 &78.9\\\hline
%      BDH+Model 1 Gold &81.8&81.1&81.4\\
%      BDH+Model 2 &77.6&76.3&76.9\\ 
%      BDH+Model TAG &78.6&76.8&\textbf{77.7}\\  \hline
%      BDH+Model 1-POS &78.8&76.7&\textbf{77.7}\\ 
%      BDH+Model TAG-POS &77.8&76.8&77.3\\ \hline
%      BDH+Model X &&&77.6\\\hline
%      Ensemble &&&\\
%      \newcite{roth2016} &79.7 &73.6& 76.5\\
%      \newcite{marcheggiani-titov:2017:EMNLP2017}&80.8&77.1& 78.9\\
%      BDH x3&77.4&76.3&76.9\\
%      BDH+Model 0 x3&78.3&77.0&77.6\\
%      BDH+Model 1 x3&79.0&76.9&77.9\\
%      BDH+Model 2 x3&&&\\
%      BDH+Model TAG x3&&&\\
%      \hline
\end{tabular}
\caption{Results on the CoNLL 2009 out-of-domain test set for English. The standard deviation in $F_1$ ranges between 0.2 and 0.35.}
\end{table}

\begin{table}
\footnotesize
\centering
\begin{tabular}{ l |ccc }
\hline
  System & P & R & F\textsubscript{1}\\\hline
%      \newcite{bjorkelund-hafdell-nugues:2009:CoNLL-2009-ST} &78.9& 74.3 &76.5\\
      \newcite{zhao-EtAl:2009:CoNLL-2009-ST2} &83.1&78.0&80.5\\
      \newcite{roth2016} &83.2&77.4&80.2\\
      \newcite{marcheggiani2017}&81.4& 79.3& 80.3\\
      \hline
      BiLSTM&79.8&78.4&79.1\\
%      BiLSTM+DOut&80.8&79.6&80.2\\
      BDH&82.0&79.7&80.8\\
%      BDH + Model 0&81.5&80.7&\textbf{81.1}\\
      BDH + Model 1 &81.9&80.2&\textbf{81.0}\\ \hline
      BDH + ELMo &83.1&82.8&82.9\\
      BDH + Model 1 + ELMo&83.1&83.0&\textbf{83.0}\\
 %     BDH + Model 1 + ELMo &&&\\
%      BDH + Model 2 &82.4&79.8&\textbf{81.1}\\
%      BDH + Model TAG &82.4&79.7&81.0\\ \hline
%      BDH + Model 1-POS &82.1&79.5&80.8\\
%      BDH + Model TAG-POS &82.2&80.0&\textbf{81.1}\\\hline
\end{tabular}
\caption{Results on the CoNLL 2009 test set for Spanish. All standard deviations in $F_1 < 0.1$.}
\label{srl-es}
%\vspace{-1.0cm}
\end{table}

\begin{figure}
\centering
\begin{tikzpicture}[scale=0.7]
\begin{axis}[
    title={},
    xlabel={},
    ylabel={F\textsubscript{1} score},
    xmin=0, xmax=120,
    ymin=86.5, ymax=91.5,
    xtick={10,30,50,70,90,110},
    xticklabels={1-10, 11-15, 16-20, 21-25, 26-30, 31-},
    ytick={87, 88,89, 90, 91},
    legend pos=north east,
    legend style={font=\footnotesize},
    ymajorgrids=true,
    grid style=dashed,
]
 
\addplot[
    color=darkgreen,
    mark=triangle,
    ]
    coordinates {(10,89.89)(30,89.84)(50,89.0)(70,87.67)(90,87.58)(110,86.97)};
    \legend{BiLSTM}
\addplot[
    color=blue,
    mark=x,
    ]
    coordinates {(10,90.2)(30,90.41)(50,89.42)(70,88.13)(90,88.33)(110,87.09)};
    \addlegendentry{BDH}
%\addplot[
%    color=green,
%    mark=triangle,
%    ]
%    coordinates {(10,90.85)(30,89.49)(50,90.25)(70,88.51)(90,88.46)(110,87.7)};
%    \addlegendentry{Model 0}
\addplot[
    color=red,
    mark=o,
    ]
    coordinates {(10,91.22)(30,90.23)(50,90.03)(70,88.29)(90,88.87)(110,87.9)};
    \addlegendentry{Model 1}
%\addplot[
%    color=black,
%    mark=square,
%    ]
%    coordinates {(10,90.16)(30,89.01)(50,89.83)(70,88.21)(90,88.63)(110,87.43)};
%    \addlegendentry{Model 2}
%\addplot[
%    color=brown,
%    mark=star,
%    ]
%    coordinates {(10,90.95)(30,90.43)(50,89.51)(70,88.42)(90,88.33)(110,87.35)};
%    \addlegendentry{Model TAG}
%\addplot[
%    color=yellow,
%    mark=star,
%    ]
%    coordinates {(10,90.65)(30,89.2)(50,89.53)(70,88.63)(90,88.45)(110,87.6)};
%    \addlegendentry{1-POS}
%\addplot[
%    color=purple,
%    ]
%    coordinates {(10,90.51)(30,90.27)(50,89.73)(70,88.49)(90,88.25)(110,87.45)};
%    \addlegendentry{TAG-POS}
\end{axis}
\end{tikzpicture}
\caption{In-domain test results by sentence length.}
\label{in-domain}
%\vspace{-0.7cm}
\end{figure}
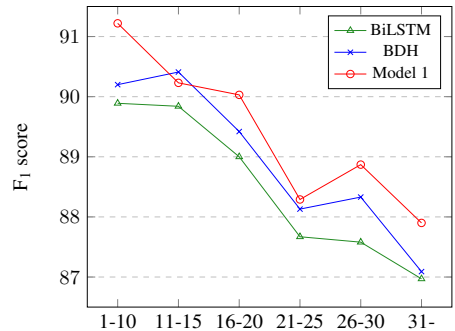

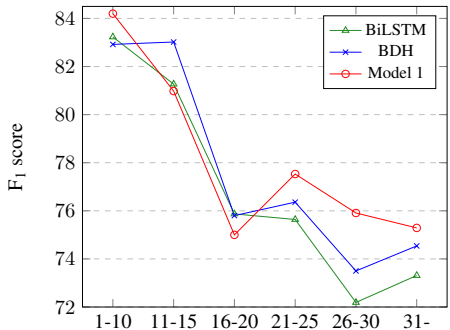
\begin{figure}
\centering
\begin{tikzpicture}[scale=0.7]
\begin{axis}[
    title={},
    xlabel={},
    ylabel={F\textsubscript{1} score},
    xmin=0, xmax=120,
    ymin=72.0, ymax=84.5,
    xtick={10,30,50,70,90,110},
    xticklabels={1-10, 11-15, 16-20, 21-25, 26-30, 31-},
    ytick={72, 74, 76, 78,80, 82, 84},
    legend pos=north east,
    legend style={font=\footnotesize},
    ymajorgrids=true,
    grid style=dashed,
]
 
\addplot[
    color=darkgreen,
    mark=triangle,
    ]
    coordinates {(10,83.23)(30,81.27)(50,75.87)(70,75.64)(90,72.19)(110,73.31)};
    \legend{BiLSTM}
\addplot[
    color=blue,
    mark=x,
    ]
    coordinates {(10,82.92)(30,83.02)(50,75.8)(70,76.36)(90,73.5)(110,74.54)};
    \addlegendentry{BDH}
\addplot[
    color=red,
    mark=o,
    ]
    coordinates {(10,84.2)(30,80.98)(50,75.0)(70,77.53)(90,75.91)(110,75.29)};
    \addlegendentry{Model 1}
\end{axis}
\end{tikzpicture}
\caption{Out-of-domain results by sentence length.}
\vspace{-0.4cm}
\label{out-of-domain}
\end{figure}

Following the analysis in \citet{roth2016}, we show plots of the BiLSTM, BDH (BiLSTM + Dropout + Highway), and Model 1 role labeling performance for sentences with varying number of words (in-domain: Fig.\ \ref{in-domain}; out-of-domain: Fig.\ \ref{out-of-domain}).
Note first that BDH outperforms the baseline BiLSTM model in a relatively uniform manner across varying sentence lengths. The benefits of Model 1 supertags, in contrast, come more from longer sentences, especially in the out-of-domain test set. 
%Following the analysis in \newcite{roth2016}, we provide a plot of the BiLSTM, BDH (BiLSTM + Dropout + Highway), and Model 1 role labeling performance for sentences with varying number of words (Fig.\ \ref{in-domain}).
%BDH outperforms the baseline BiLSTM model in a relatively uniform manner across varying sentence lengths. On the other hand, the gains in Model 1 come more from longer sentences.
This implies that the supertag model is robust to the sentence length, probably because supertags encode relations between words that are linearly distant in the sentence, information that a simple BiLSTM is unlikely to recover.

\begin{table*}
\footnotesize
\centering
\begin{tabular}{c|ccc|ccc|ccc|ccc}
\hline
    &\multicolumn{3}{c|}{V \!/\! A0} &\multicolumn{3}{c|}{V \!/\! A1} & \multicolumn{3}{c|}{ V \!/\! A2} &\multicolumn{3}{c}{V \!/\! AM} \\
  Model & P&R&F&P&R&F&P&R&F&P&R&F\\\hline
  Mate-tools & 91.2&87.4&89.3 &91.0&90.8&90.9&82.8&76.9&79.7&79.3&74.4&76.8\\
  Path-LSTM & 90.8&89.2&90.0&91.0 &91.9&91.4&84.3&76.9&80.4&82.2&72.4&77.0\\\hline
 BiLSTM & 91.1&89.7&90.4&92.1&90.9&91.5&84.0&75.0&79.2&77.7&76.9&77.3\\
  BDH & 90.9&90.8&90.9&91.5&92.4&92.0&80.3&76.1&78.1&79.6&79.1&79.3\\
  Model 0 &92.3&92.2&\textbf{92.3}&93.4&92.7&\textbf{93.0}&81.9&77.8&79.8&79.1&79.5&79.3\\ 
  Model 1 &92.5&91.6&92.0&93.0&92.8&92.9&80.9&80.3&\textbf{80.6}&80.1&78.6&\textbf{79.4}\\
  Model 2 &91.9&90.1&91.0&92.5&92.4&92.4&79.2&77.8&78.5&79.9&78.2&79.1\\ 
  TAG & 91.7&89.9&90.8&92.5&93.3&92.9&82.1&77.3&79.6&80.4&78.3&79.3\\
%  1-POS & 92.5&90.3&91.4&92.7&93.2&92.9&82.4&75.1&78.6&80.4&78.9&\textbf{79.6}\\
%  TAG-POS & 92.1&90.1&91.1&92.8&92.9&92.8&80.3&78.2&79.3&80.5&78.7&\textbf{79.6}\\
  
  \hline
  \hline
    &\multicolumn{3}{c|}{N \!/\! A0} &\multicolumn{3}{c|}{N \!/\! A1} & \multicolumn{3}{c|}{ N \!/\! A2} &\multicolumn{3}{c}{N \!/\! AM} \\
  Model & P&R&F&P&R&F&P&R&F&P&R&F\\\hline
  Mate-Tools & 86.1&74.9&80.2&84.9&82.2&83.5&81.4&74.7&77.9&78.6&72.0&75.2\\
  Path-LSTM & 86.9&78.2&82.3&87.5&84.4&\textbf{85.9}&82.4&76.8&79.5&79.5&69.2&74.0\\\hline
  BiLSTM & 85.1&79.5&82.2&85.8&83.4&84.6&81.0&76.4&78.7&72.8&71.4&72.1\\
  BDH & 83.9&80.1&82.0&84.8&86.1&85.5&80.6&77.0&78.8&71.3&77.9&74.5\\
  Model 0 &87.2&77.6&82.1&86.2&85.4&85.8&79.9&79.2&79.5&69.4&79.4&74.0 \\
  Model 1 &84.1&80.7&82.4&85.2&86.0&85.6&79.6&79.6&\textbf{79.6}&75.2&76.3&\textbf{75.8}\\
  Model 2 & 86.0&79.5&\textbf{82.6}&85.4&85.5&85.5&80.5&77.3&78.9&73.3&76.1&74.6\\
  TAG & 83.9&79.8&81.8&84.9&86.3&85.6&81.5&75.8&78.6&72.3&72.9&72.6\\ \hline
%  1-POS & 84.3&79.6&81.9&85.1&85.7&85.4&82.3&75.5&78.8&73.0&77.4&75.1\\ 
%  TAG-POS & 86.1&77.5&81.6&85.2&86.4&\textbf{85.8}&79.8&78.8&79.3&71.3&75.9&73.5\\  \hline
\end{tabular}
\caption{English in-domain test results by predicate category and role label. The mate-tools \cite{bjorkelund-hafdell-nugues:2009:CoNLL-2009-ST} and Path-LSTM results are taken from \newcite{roth2016}.}
\label{wordrole}
\end{table*}

Table \ref{wordrole} reports SRL results broken down by predicate category (V: Verb, Propbank; N: Noun, Nombank) and semantic role.
We can observe  that the various supertag models differ in their performance for different predicate-role pairs, suggesting that different  kinds of linguistic information are relevant for identifying the different roles.  Overall, Model 1 supertags achieve the most consistent improvements over BiLSTM and BiLSTM + Dropout + Highway (BDH) in V \!/\! A0,  V \!/\! A1,  V \!/\! A2, V \!/\! AM, N \!/\! A2, and N \!/\! AM.
Moreover, Model 1 even improves on Path-LSTM \cite{roth2016} by large margins in V \!/\! A0,  V \!/\! A1, V \!/\! AM, and N \!/\! AM, 
% implying that it serves as an effective middle ground between the syntax-agnostic  and full parsing approaches.
even though the Path-LSTM model has the benefit of using the complete dependency path between each word and its head. This shows that supertags can be even more effective for SRL than more granular syntactic information--even quite simple supertags, like Model 0, which encode only the dependency arc between a word and its head.

\vspace{-2mm}
\section{Conclusion and Future Work}
We presented state-of-the-art SRL systems on the CoNLL 2009 English and Spanish data that make crucial use of dependency-based supertags. We showed that supertagging serves as an effective middle ground between syntax-agnostic approaches and full parse-based approaches for dependency-based semantic role labeling.
Supertags give useful syntactic information for SRL and allow us to build an SRL system that does not depend on a complex architecture.  We have also seen that the choice of the linguistic content of a supertag makes a significant difference in its utility for SRL.
In this work, all models are developed independently for English and Spanish. However, sharing some part of SRL models could improve performance \cite{mulcaire-swayamdipta-smith:2018:Short, mulcaire2019-naacl}. In future work, we will explore crosslingual transfer for supertagging and semantic role labeling.

%We predicted supertags and semantic role labels independently for English and Spanish.
%However, sharing some part of a supertagger and role labeler across languages could improve supertagging and SRL performance \cite{mulcaire-swayamdipta-smith:2018:Short}.
%We will explore such possibilities in future.
%In particular, if a pair of languages has morphological similarity such as Spanish and Catalan, we can jointly learn character-level representation.
%In future work, we will explore multilingual transfer learning for supertagging and semantic role labeling.

\section*{Acknowledgments}
The authors thank Diego Marcheggiani for assistance in implementing SRL models and Diego Marcheggiani and the anonymous reviewers for their helpful feedback. This work was funded in part by the Funai Overseas Scholarship to JK.

\bibliography{naaclhlt2019}
\bibliographystyle{acl_natbib}

\newpage
\appendix
%\newpage
\section{Hyperparameters}
\label{hyper}
\begin{table}
\small
\centering
\begin{tabular}{ |l r|}
\hline
$d_w$ (English word embeddings) & 100\\
$d_w$ (Spanish word embeddings) & 300\\
$d_{pos}$ (POS embeddings) & 100\\
Char-CNN window size &3\\
Char-CNN \# filters &30\\
Char-CNN character embedding size & 30\\
$d_h$ (LSTM hidden states) & 512\\
$k$ (BiLSTM depth) & 4\\
LSTM dropout rate & 0.5\\
Recurrent dropout rate & 0.5\\
Batch Size & 100\\
Adam \cite{Kingma2015} lrate& 0.01\\
Adam $\beta_1$& 0.9\\
Adam $\beta_2$& 0.999\\
\hline
\end{tabular}
\caption{Supertagging Hyperparameters.}
\label{stag-hyp}
\end{table}
All of our models are implemented in TensorFlow \cite{tensorflow2015-whitepaper}.

\paragraph{Supertagging}
We follow the hyperparameters chosen in \newcite{kasai-EtAl:2018:N18-1}.
Specifically, we list the hyperparameters in Table \ref{stag-hyp} for completeness and easy replication.

\paragraph{SRL}
\begin{table}
\small
\centering
\begin{tabular}{ |l r|}
\hline
$d_w$ (English word embeddings) & 100\\
$d_w$ (Spanish word embeddings) & 300\\
$d_{pos}$ (POS embeddings) & 16\\
$d_l$ (lemma embeddings) & 100\\
$d_s$ (supertag embeddings) & 50\\
$d_h$ (LSTM hidden states) & 512\\
$d_r$ (role representation) & 128\\
$d'_l$ (output lemma representation) & 128\\
$k$ (BiLSTM depth) & 4\\
$\alpha$ (word dropout) & .25\\
LSTM dropout rate & 0.5\\
Batch Size & 100\\
Adam lrate& 0.01\\
Adam $\beta_1$& 0.9\\
Adam $\beta_2$& 0.999\\
\hline
\end{tabular}
\caption{SRL Hyperparameters}
\label{srl-hyp}
\end{table}
We follow the hyperparameters of \citet{marcheggiani2017} and add highway connections \cite{he2017} and LSTM dropout.
Concretely, we use the hyperparameters shown in Table \ref{srl-hyp}.

\paragraph{Contextual Representations}
\begin{table}
\small
\centering
\begin{tabular}{ |l p{0.3\linewidth}|}
\hline
\multicolumn{2}{|c|}{Character CNNs}\\
Char embedding size & 16\\
(\# Window Size, \# Filters) & (1, 32), (2, 32), (3, 68), (4, 128), (5, 256), 6, 512), (7, 1024)\\ 
Activation & Relu\\
\hline
\multicolumn{2}{|c|}{Word-level LSTM}\\
LSTM size & 2048\\
\# LSTM layers & 2\\
LSTM projection size & 256\\
Use skip connections & Yes\\
Inter-layer dropout rate& 0.1\\
\hline
\multicolumn{2}{|c|}{Training}\\
Batch size & 128\\
Unroll steps (Window Size) & 20\\
\# Negative samples & 64\\
\# Epochs & 10\\
Adagrad \cite{journals/jmlr/DuchiHS11} lrate& 0.2\\
Adagrad initial accumulator value& 1.0 \\
\hline
\end{tabular}
\caption{Spanish Language Model Hyperparameters.}
\label{lm-hyp}
\end{table}
For English, we use the pretrained ELMo model available at \url{https://tfhub.dev/google/elmo/2}.
For Spanish, we use a multilingual fork \cite{mulcaire2019-naacl}\footnote{\url{https://github.com/pmulcaire/rosita/}} of the AllenNLP library \cite{Gardner2017AllenNLP}, and train a language model on the pre-segmented Spanish data provided by \citet{Conll2017data}.\footnote{\url{https://lindat.mff.cuni.cz/repository/xmlui/handle/11234/1-1989}}
We follow the hyperparameters chosen in \citet{mulcaire2019-naacl} (Table \ref{lm-hyp}), and randomly sample 50 million tokens from the Spanish data for training.

\section{Supplementary Analysis}
\begin{figure*}
\centering
\includegraphics[width=1.0\textwidth]{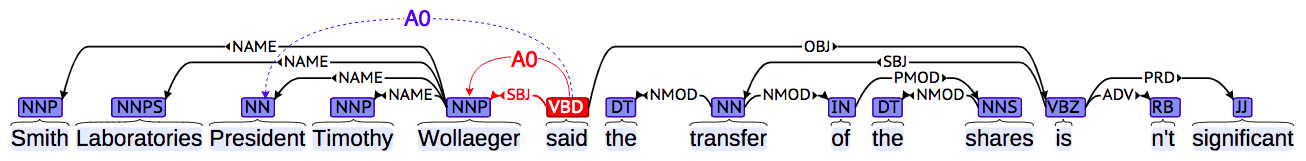}
\caption{V \!/\! A0 case where BDH assigns A0 to \textit{President} (blue arc) while Model 1 correctly assigns A0 to \textit{Wollaeger} (red arc). The predicted Model 1 supertags for \textit{President} and \textit{Wollaeger} are \textbf{NAME/R} and \textbf{SBJ/R+L} respectively.}
\label{V-A0}
\end{figure*}

\begin{figure*}
\centering
\includegraphics[width=1.0\textwidth]{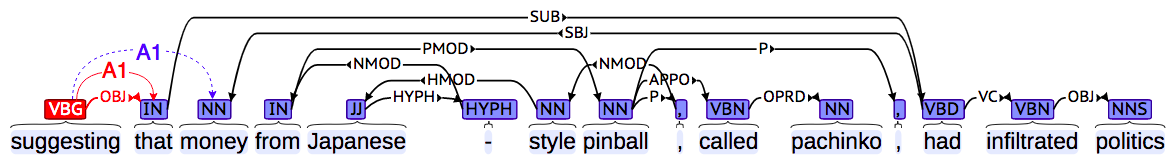}
\label{V-A1}
\caption{V \!/\! A1 where BDH assigns A1 to \textit{money} (blue arc) while Model 1 correctly assigns A1 to \textit{that} (red arc). The predicted Model 1 supertags for \textit{that} and \textit{money} are \textbf{OBJ/L+R} and \textbf{SBJ/R+R} respectively.}
\end{figure*}

\begin{figure*}
\centering
\includegraphics[width=1.0\textwidth]{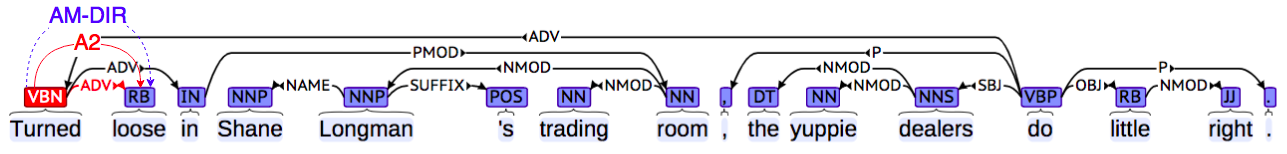}
\caption{V \!/\! A2 case where BDH assigns AM-DIR to \textit{loose} (blue arc) while Model 1 correctly assign A2 (red arc). The predicted supertag for \textit{loose} is \textbf{PRD/L} (predicative complement). Notice that the ``PRT" (particle) or ``DIR'' (adverbial of direction) feature is not predicted that could have misled the labeling. Interestingly, the gold parse and gold POS tag for \textit{loose} treat it as an adverbial modifier to \textit{turned}.}
\label{V-A2}
\end{figure*}

\begin{figure*}
\centering
\includegraphics[width=1.0\textwidth]{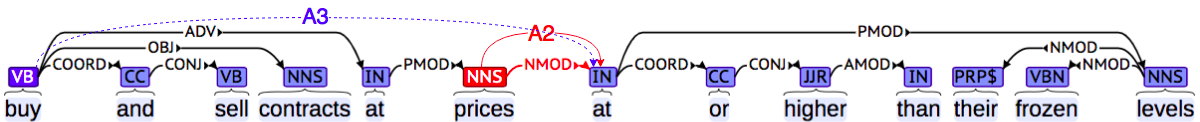}
\caption{N \!/\! A2 case where BDH assigns A3 for the predicate \textit{buy} to \textit{at} (blue arc) while Model 1 correctly assigns A2 for the predicate \textit{prices} (red arc). The predicted Model 1 supertag for \textit{at} was \textbf{NMOD/L+R}, correctly resolving the PP attachment ambiguity.}
\label{N-A2}
\end{figure*}

We show examples from the dev set in Figures \ref{V-A0}-\ref{N-A2} where a model without supertags mislabels (dashed blue arcs) and Model 1 (red arcs) correctly labels.
In all those cases, it is clear that the predicted supertags are playing a crucial role in guiding role labeling. 

\newpage


\end{document}